\pdfoutput=1

\documentclass[11pt]{article}

\usepackage{acl}

\usepackage{times}
\usepackage{latexsym}

\usepackage[T1]{fontenc}

\usepackage[utf8]{inputenc}

\usepackage{microtype}

\usepackage{algorithm}
\usepackage{algorithmic}
\usepackage{amssymb}
\usepackage{helvet}  
\usepackage{courier}  
\usepackage{newfloat}
\usepackage{listings}
\usepackage{multirow}
\usepackage{textcomp}
\usepackage{stfloats}
\usepackage{url}
\usepackage{verbatim}
\usepackage{graphicx}
\usepackage{amsmath}
\usepackage{booktabs}
\usepackage{pifont}
\usepackage{color}

\newcommand{\model}{KECP}

\title{KECP: Knowledge Enhanced Contrastive Prompting for Few-shot Extractive Question Answering}

\author{Jianing Wang$^1$\thanks{\ \ \ J. Wang and C. Wang contributed equally to this work.}, Chengyu Wang$^2$\footnotemark[1], Minghui Qiu$^2$, Qiuhui Shi$^3$, Hongbin Wang$^3$,\\ {\bf Jun Huang$^2$, Ming Gao$^1$\thanks{\ \ \ Corresponding author.}}\\
  $^1$ School of Data Science and Engineering, East China Normal University \\
  $^2$ Alibaba Group $^3$ Ant Group\\
  \texttt{lygwjn@gmail.com, mgao@dase.ecnu.edu.cn}\\
  \texttt{\{chengyu.wcy,minghui.qmh,huangjun.hj\}@alibaba-inc.com}\\
  \texttt{\{hongbin.whb,qiuhui.sqh\}@antgroup.com}
  }

\begin{document}
\maketitle
\begin{abstract}
Extractive Question Answering (EQA) is one of the most important tasks in Machine Reading Comprehension (MRC), which can be solved by fine-tuning the span selecting heads of Pre-trained Language Models (PLMs). However, most existing approaches for MRC may perform poorly in the few-shot learning scenario. To solve this issue, we propose a novel framework named \textbf{K}nowledge \textbf{E}nhanced \textbf{C}ontrastive \textbf{P}rompt-tuning (\emph{\model}). Instead of adding pointer heads to PLMs, we introduce a seminal paradigm for EQA that transform the task into a non-autoregressive Masked Language Modeling (MLM) generation problem.  Simultaneously, rich semantics from the external knowledge base (KB) and the passage context are support for enhancing the representations of the query. In addition, to boost the performance of PLMs, we jointly train the model by the MLM and contrastive learning objectives. Experiments on multiple benchmarks demonstrate that our method consistently outperforms 
state-of-the-art approaches in few-shot settings by a large margin. 
\footnote{All datasets are publicly available. Source codes will be released in EasyNLP~\cite{DBLP:journals/corr/abs-2205-00258}. URL:~\url{https://github.com/alibaba/EasyNLP}}

\end{abstract}

\section{Introduction}

\begin{figure}[t]
\centering
\includegraphics[width=\columnwidth]{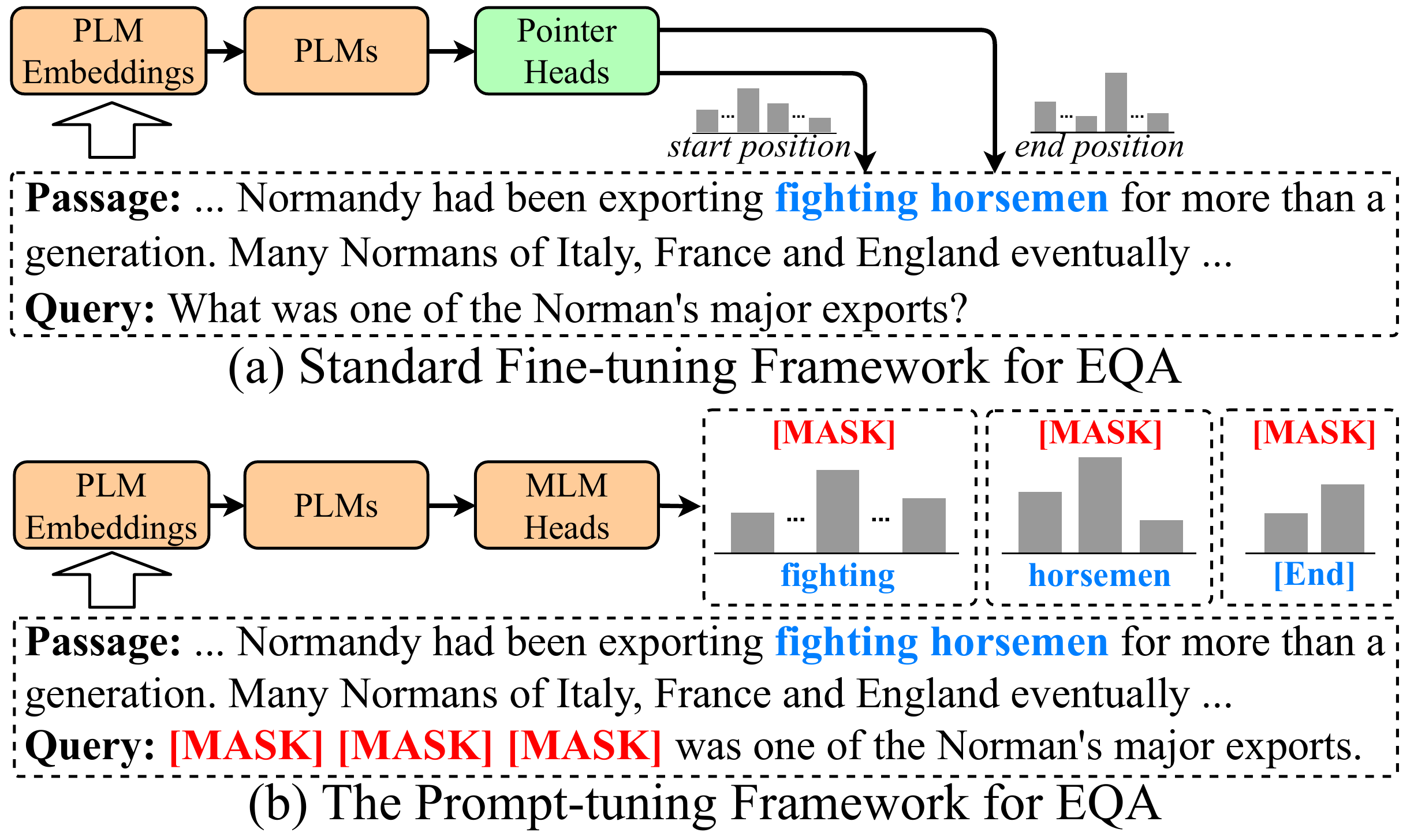} 
\caption{The comparison of the standard fine-tuning and prompt-tuning framework. The blocks in orange and green denote the modules of PLMs and newly initialized modules, respectively. 
(Best viewed in color.)}
\label{fig1}
\end{figure}

Span-based Extractive Question Answering (EQA) is one of the most challenging tasks of Machine Reading Comprehension (MRC). 
A majority of recent approaches~\cite{Wang2019Explicit,Yang2019Enhancing,Dai2021Incorporating} add pointer heads~\cite{Vinyals15Pointer} to Pre-trained Language Models (PLMs) to predict the start and the end positions of the answer span (shown in Figure~\ref{fig1}(a)).
Yet, these conventional fine-tuning frameworks heavily depend on the time-consuming and labor-intensive process of data annotation. Additionally, there is a large gap between the pre-training objective of Masked Language Modeling (MLM) (i.e.,~predicting the distribution over the entire vocabularies) and the fine-tuning objective of span selection (i.e.,~predicting the distribution of positions), which hinders the transfer and adaptation of knowledge in PLMs to downstream MRC tasks~\cite{Brown2020Language}.
A straightforward approach is to integrate the span selection process into pre-training~\cite{Ram2021Few}. However, it may cost a lot of computational resources during pre-training.

Recently, a branch of prompt-based fine-tuning paradigm (i.e. prompt-tuning) arises to transform the downstream tasks into the cloze-style problem~\cite{Timo2021Exploiting,Xu2021PTR,Xiang2021Prefix,Gao2021Making,Liu21PTuningv2}. To specify, task-specific prompt templates with \texttt{[MASK]} tokens are added to input texts (\texttt{[MASK]} denotes the masked language token in PLMs). The results of the masked positions generated by the MLM head are used for the prediction\footnote{For example, in sentiment analysis, a prompt template~(e.g., ``It was \texttt{[MASK]}.'') is added to the review text (e.g., ``This dish is very attractive.''). We can obtain the result tokens of masked position for label prediction (e.g., ``delicious'' for the positive label and ``unappetizing'' for the negative label).}.
By prompt-tuning, we can use few training samples to fast adapt the prior knowledge in PLMs to downstream tasks.
A natural idea is that we can transform EQA into the MLM task by adding a series of masked language tokens. As shown in Figure \ref{fig1}(b),
the query is transformed into a prompt template containing multiple \texttt{[MASK]} tokens, which can be directly used for the answer tokens prediction. However, we observe that two new issues for vanilla PLMs: 1) the MLM head, which is based on single-token non-autoregressive prediction, has a poor inference ability to understand the task paradigm of EQA
; 2) there are many confusing span texts in the passage have similar semantics to the correct answer, which can unavoidably make the model produce negative answers.
Therefore, a natural question arises: \textit{how to employ prompt-tuning over PLMs for EQA to achieve high performance in the few-shot learning setting?} 

In this work, we introduce \textit{\model}, a novel \textbf{K}nowledge \textbf{E}nhanced \textbf{C}ontrastive \textbf{P}rompting framework for the EQA task. 
We view EQA as an MLM generation task that transform the query to a prompt with multiple masked language tokens.
In order to improve the inference ability, for each given example, we inject related knowledge base (KB) embeddings into context embeddings of the PLM, and enrich the representations of selected tokens in the query prompt.
To make PLMs better understand the span prediction task, we further propose a novel span-level contrastive learning objective to boost the PLM to distinguish the correct answer with the negatives with similar semantics.
During the inference time, we implement a highly-efficient model-free prefix-tree decoder and generate answers by beam search.
In the experiments, we evaluate our proposed framework over seven EQA benchmarks in the few-shot scenario.
The results show that our method consistently outperforms
state-of-the-art approaches by a large margin. Specifically, we achieve a 75.45\% F1 value on SQuAD2.0 with only 16 training examples.

To sum up, we make the following contributions:
\begin{itemize}
\item We propose a novel \textit{\model} framework for few-shot EQA task based on prompt-tuning. 

\item In \textit{\model}, EQA is transformed into the MLM generation problem, which alleviates model over-fitting and bridges the gap between pre-training and fine-tuning. We further employ knowledge bases to enhance the token representations and design a novel contrastive learning task for better performance.

\item Experiments show that \textit{\model} outperforms all the baselines in few-shot scenarios for EQA.
\end{itemize}

\section{Related Work}
In this section, we summarize the related work on EQA and prompt-tuning for PLMs.

\subsection{Extractive Question Answering}
EQA is one of the most challenging MRC tasks, which aims to find the correct answer span from a passage based on a query. A variety of  benchmark tasks on EQA have been released and attracted great interest~\cite{Rajpurkar2016SQuAD,Fisch2019MRQA,Rajpurkar2018Know, Lai2017RACE, trischler2017newsqa, Levy2017Zero, Joshi17TriviaQA}. Early works utilize attention mechanism to capture rich interaction information between the passage and the query~\cite{wang2017gated,Wang2017Machine}. Recently, benefited from the powerful modeling abilities of PLMs, such as GPT~\cite{Brown2020Language}, BERT~\cite{Devlin2019BERT}, RoBERTa~\cite{Liu19RoBERTa} and SpanBERT~\cite{Joshi20SpanBERT}, etc., we have witnessed the qualitative improvement of MRC based on fine-tuning PLMs.
However, this standard fine-tuning paradigm may cause over-fitting in the few-shot settings. To solve the problem, ~\cite{Ram2021Few} propose Splinter for few-shot EQA by pre-training over the span selection task, but it costs a lot of time and computational resources to pre-train these PLMs. On the contrary, we leverage prompt-tuning for few-shot EQA without any additional pre-training steps.

\subsection{Prompt-tuning for PLMs}


Prompt-tuning is one of the flourishing research in the past two years. GPT-3~\cite{Brown2020Language} enables few/zero-shot learning for various NLP tasks without fine-tuning, which relies on handcraft prompts and achieves outstanding performance. To facilitate automatic prompt construction, AutoPrompt~\cite{Shin2020AutoPrompt} and LM-BFF~\cite{Gao2021Making} automatically generate discrete prompt tokens from texts. Recently, a series of methods learn continuous prompt embeddings with differentiable parameters for natural language understanding and text generation task, such as Prefix-tuning~\cite{Li2021Prefix}, P-tuning V2~\cite{Liu21PTuningv2}, PTR~\cite{Xu2021PTR}, and many others~\cite{Lester2021The, Zou2021Controllable, Li2021Prefix, Qin2021Learning, Schick2021It}. 
Different from previous work~\cite{Ram2021Few}, our prompting-based framework mainly focuses on EQA, which is the novel exploration for this challenging task in few-shot learning setting.

\begin{figure*}[t]
\centering
\includegraphics[width=\textwidth]{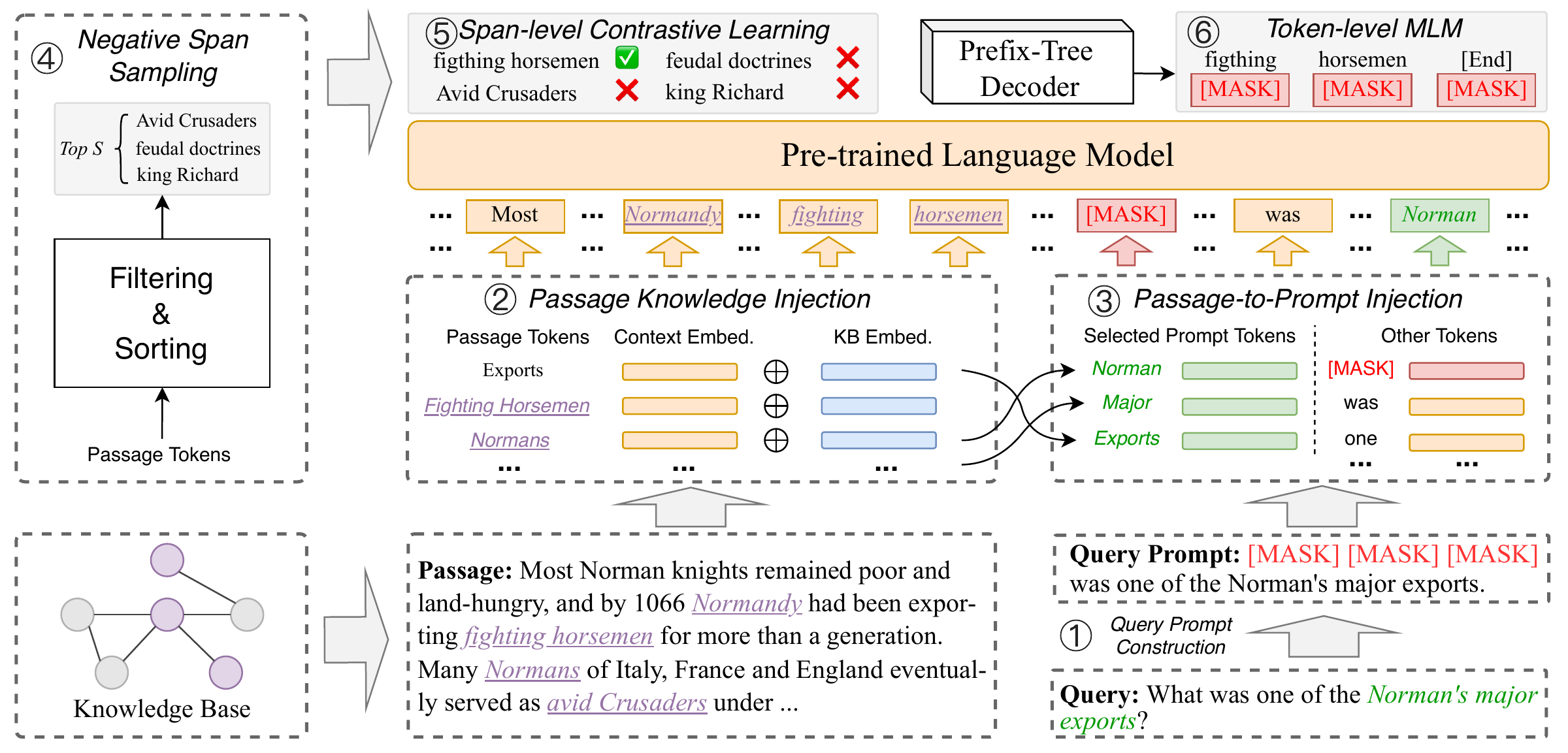} 
\caption{The \textit{\model} framework. Given a passage and a query, we first construct the query prompt by heuristic rules (\ding{172}). Next, we capture the knowledge both from passage text and external KB to enhance the representations of selected prompt tokens (\ding{173} \ding{174}). To improve the accuracy of answer prediction, we sample negative span texts with similar and confused semantics (\ding{175}), and train the model with contrastive learning (\ding{176}). During the inference stage, the answer span text can be generated by MLM and a model-free prefix-tree decoder (\ding{177}). (Best viewed in color).}
\vspace{-.5em}
\label{fig2}
\end{figure*}

\section{The~\textit{\model} Framework}
In this section, we formally present our task and the techniques of the~\textit{\model} framework in detail. The overview of~\textit{\model} is shown in Figure \ref{fig2}.

\subsection{Task Overview}

Given a passage $P=p_1, \cdots, p_n$ and the corresponding query $Q=q_1, \cdots, q_m$, the goal is to find a sub-string of the passage as the answer $Y=p_k, \cdots, p_{l}$, where $n, m$ are the lengths of the passage, the query, respectively. $p_{i}$ ($i=1,\cdots,n$) and $q_{j}$ ($j=1, \cdots, m$) refer to the tokens in $P$ and $Q$, respectively. $k, l$ denotes the start and end position of the passage, $1\leq k\leq l \leq n$.
Rather than predict the start and the end positions of the answer span, we view the EQA task as a non-autoregressive MLM generation problem.
In the following, we will provide the detailed techniques of the \emph{\model} framework.


\subsection{Query Prompt Construction}

Since we transform the conventional span selection problem into the MLM generation problem, we need to construct prompt templates for each passage-query pair. In contrast to previous approaches~\cite{Brown2020Language,Gao2021Making} which generate templates by handcrafting or neural networks, we find that the query $Q$ in EQA tasks naturally provides hints for prompt construction. Specifically, we design a template mapping $\mathcal{T}$ based on several heuristic rules (please refer to Appendix~\ref{appendix:query_prompt} for more details). For example, the query ``{What was one of the Norman's major exports?}'' can be transformed into a template: ``\texttt{[MASK][MASK][MASK]} {was one of the Norman's major exports}''. If a sentence does not match any of these rules, multiple \texttt{[MASK]} tokens will be directly added to the end of the query.
The number of \texttt{[MASK]} tokens in prompts is regarded as a pre-defined hyper-parameter denotes as $l_{mask}$.

Let $Q_{prompt}=q'_1, q'_2,\cdots, q'_{m'}$ denote a query prompt where $q'_i$ is a dispersed prompt token, $m'$ is the length of query prompt. We concatenate the query prompt $Q_{prompt}$ and the passage text $P$ with some special tokens as input $x_{input}$:
\begin{equation}
x_{input} = \texttt{[CLS]} Q_{prompt} \texttt{[SEP]} P \texttt{[SEP]},
~\label{eqn:input}
\end{equation}
where \texttt{[CLS]} and \texttt{[SEP]} are two special tokens that represent the start and separate token in PLMs. 

\subsection{Knowledge-aware Prompt Encoder (KPE)}
\label{sec-kpe}

As mentioned above, to remedy the dilemma that vanilla MLM has poor abilities of model inference, empirical evidence suggests that we can introduce the KB to assist boosting PLMs. For example, when we ask the question ``What was one of the Norman's major exports?'', we expect the model to capture more semantics information of the selected tokens ``Norman's major exports'', which is the imperative component for model inference.

To achieve this goal, inspired by \citet{Xiao2021GPT} where pseudo tokens are added to the input with continuous prompt embeddings,
we propose the \textit{Knowledge-aware Prompt Encoder} (KPE) to aggregate the multiple-resource knowledge to the input embeddings of the query prompt. It consists of two main steps: \textit{Passage Knowledge Injection} (PKI) and \textit{Passage-to-Prompt Injection} (PPI), where the first aims to generate knowledge-enhanced representations from passage context and KB, while the second is to flow these representations to the selected tokens of query prompts. 

\subsubsection{Passage Knowledge Injection (PKI)}
For knowledge injection, we first introduce two embedding mappings $\mathcal{E}_{wr}(\cdot)$ and $\mathcal{E}_{kn}(\cdot)$, where $\mathcal{E}_{wr}(\cdot)$ aims to map the input token to the word embeddings from the PLM embedding table, $\mathcal{E}_{kn}(\cdot)$ denotes to map the input token to the KB embeddings pre-trained by the ConVE~\cite{Tim18Convolutional} algorithm based on WikiData5M~\cite{Wang21KEPLER}~\footnote{URL: \url{https://deepgraphlearning.github.io/project/wikidata5m}.}.

In the beginning, all the tokens in $x_{input}$ are encoded into word embeddings $\mathbf{x}$. Hence, we can obtain the embeddings of query prompt and passage, denote as $\mathbf{Q}=\mathcal{E}_{wr}(Q_{prompt}) \in \mathbb{R}^{m'\times h}$ and $\mathbf{P}=\mathcal{E}_{wr}(P) \in \mathbb{R}^{n\times h}$. 
Additionally, for each token $p_i\in P$, we retrieve the entities from the KB that have the same lemma with $p_i$, and the averaged entity embeddings are stored as their KB embeddings. Formally, we generate the KB embeddings $\mathbf{p}_i^{kn}$ of the passage token $p_i$:
\begin{equation}
\begin{aligned}
\mathbf{p}_i^{kn} = Mean(\mathbf{e}_{j}|lem(p_i) = lem(e_j)),
\label{eqn:lemma}
\end{aligned}
\end{equation}
where $lem$ is the lemmatization operator~\cite{Dai2021Incorporating}, $\mathbf{e}_{j}=\mathcal{E}_{kn}(e_j)$. We then directly combine word embeddings and KB embeddings by $\mathbf{g}_{i} = \mathbf{p}_i + \mathbf{p}_i^{kn}$,
where $\mathbf{p}_i$ is the word embeddings of $i$-th token in the passage text. $\mathbf{g}_i\in\mathbb{R}^{h}$ is the embeddings with knowledge injected. 
Finally, we obtain knowledge-enhanced representations denoted as $\mathbf{G} = \mathbf{g}_1\mathbf{g}_2, \cdots, \mathbf{g}_{n}$, where $\mathbf{G}\in\mathbb{R}^{n\times h}$.

\subsubsection{Passage-to-Prompt Injection (PPI)}

The goal of PPI is to enhance the representations $\mathbf{Q}$ of selected prompt tokens by the interaction between the query and the passage representations. 
As discovered by~\cite{zhang2021drop}, injecting too much background knowledge may harm the performance of downstream tasks, hence we only inject knowledge to the representations of part of the prompt tokens.
To be more specific, given $r (<m')$ selected prompt tokens $q_{j}^{sp}\in Q_{prompt}$, we create the corresponding embeddings $\mathbf{q}^{sp}\in\mathbb{R}^{r\times h}$ by looking up the embeddings from $\mathbf{Q}$. 
For each prompt token, we leverage self-attention to obtain the soft embeddings $\mathbf{v}^{sp}\in\mathbb{R}^{r\times h}$:
\begin{equation}
\begin{aligned}
\mathbf{v}^{sp} = \text{SoftMax}(\mathbf{q}^{sp}\mathbf{W}_{\alpha}\mathbf{G}^{\mathsf{T}}/\sqrt{d})\mathbf{G},
\label{eqn:attention-1}
\end{aligned}
\end{equation}
where $\mathbf{W}_{\alpha}\in\mathbb{R}^{h\times h}$ is the trainable matrix. $d$ denotes the scale value. 
We add residual connection to $\mathbf{v}^{sp}$ and $\mathbf{q}^{sp}$ by linear combination as $\mathbf{u}^{sp} = \mathbf{v}^{sp} + \mathbf{q}^{sp}$, where $\mathbf{u}^{sp}$ denotes the enhanced representations of selected prompt tokens.

Finally, we only replace the original word embeddings $\mathbf{x}$ of selected prompt tokens $\mathbf{q}^{sp}$ with $\mathbf{u}^{sp}$ in the PLM's embeddings layer. 
To this end, we use very few parameters to implement the rich knowledge injection, which alleviate over-fitting during few-shot learning.



\subsection{Span-level Contrastive Learning (SCL)}
As mentioned above, many negative span texts in the passage have similar and confusing semantics with the correct answer. This may cause the PLM to generate wrong results. For example, given the passage ``Google News releases that Apple founder Steve Jobs will speak about the new iPhone 4 product at a press conference in 2014.'' and the query ``Which company makes iPhone 4?''. The model is inevitably confused by some similar entities. For examples, ``Google'' is also a company name but is insight of the entity ``Apple'' in the sentence, and ``Steve Jobs'' is not a company name although it is as expected from the answer.

Inspired by contrastive learning~\cite{Chen2020a}, we can distinguish between the positive and negative predictions and alleviate this confusion problem. 
Specifically, we firstly obtain a series of span texts by the slide window, suppose as $Y'_{i}=p_{k'_i}\cdots p_{l'_i}$, where $k'_i$ and $l'_i$ denote the start and the end positions of the $i$-th span. Then, we filter out some negative spans that have similar semantics with the correct answer $Y$. In detail, we follow SpanBERT~\cite{Joshi20SpanBERT} to represent each span by the span boundary. The embeddings that we choose are the knowledge-enhanced representations $\mathbf{G}$ in Section~\ref{sec-kpe}, which consists of rich context and knowledge semantics. For each positive-negative pair $(Y, Y'_i)$, we compute the similarity score and the candidate intervals with top-$S$ similarity scores are selected as the negative answers, which can be viewed as the semantically confusion w.r.t. the correct answer.
For the $i$-th negative answer $Y'_{i}$, we have:
\begin{equation}
\begin{aligned}
\mathcal{Z}'_i= \sum_{j}\Pr(Y'_{ij}|P, Q;\Theta),
\label{eqn:score}
\end{aligned}
\end{equation}
where $\Pr$ denotes the prediction function of the MLM head. $Y'_{ij}$ denotes the $j$-th token in the corresponding span. We can also calculate the score $\mathcal{Z}$ of the ground truth in the same manner. Hence, for each training sample, the objective of the span-level contrastive learning can be formulated as:
\begin{equation}
\begin{aligned}
\mathcal{L}_{SCL} = - \frac{1}{S + 1}\log[\frac{\exp\{\mathcal{Z}\}}{\exp\{\mathcal{Z}\} + \sum_{i=1}^{S} \exp\{\mathcal{Z}'_i\}}],
\label{eqn:loss-csc}
\end{aligned}
\end{equation}
Finally, the total loss function is written as follows:
\begin{equation}
\mathcal{L} = \mathcal{L}_{MLM} + \lambda\mathcal{L}_{SCL} + \gamma||\Theta||,
\label{eqn:loss}
\end{equation}
where $\mathcal{L}_{MLM}$ denotes the training objective of token-level MLM. $\Theta$ denotes the model parameters. $\lambda, \gamma\in[0,1]$ are the balancing hyper-parameter and the regularization hyper-parameter, respectively.

\subsection{Model-free Prefix-tree Decoder}
Different from conventional text generation, we should guarantee that the generated answer must be the \textbf{sub-string} in the passage text. In other words, the searching space of each position is constrained by the prefix token. For example, in Figure \ref{fig2}, if the prediction of the first \texttt{[MASK]} token in $Q_{prompt}$ is ``fighting'', the searching space of the second token shrinks down to ``\{horsemen, \texttt{[END]}\}'', where \texttt{[END]} is the special token as the answer terminator. We implement a simple model-free prefix-tree (i.e. trie-tree) decoder without any parameters, which is a highly-efficient data structure that preserves the dependency of each passage token. At each \texttt{[MASK]} position, we use beam search algorithm to select top-$S$ results. The predicted text of the masked positions with highest score calculated by Eq.~(\ref{eqn:score}) is selected as the final answer.

\section{Experiments}

In this section, we conduct extensive experiments to evaluate the performance of our framework.

\begin{table*}[t]
\centering
\begin{small}
\resizebox{\linewidth}{!}{
\begin{tabular}{c|c|c|c|c|c|c|c|c|c}
\toprule
\bf \multirow{2}*{Paradigm} & \bf \multirow{2}*{Methods} & \bf \multirow{2}*{Use KB} & \multicolumn{7}{c}{\bf EQA Datasets}  \\
~ & ~ & ~ & \bf SQuAD2.0 & \bf SQuAD1.1 & \bf NewsQA & \bf TriviaQA & \bf SearchQA & \bf HotpotQA & \bf NQ. 	 \\
\midrule
\multirow{4}*{FT} & RoBERTa & No & 9.55\%\scriptsize $+$1.9 & 12.50\%\scriptsize $+$2.7 & 6.24\%\scriptsize $+$0.8 & 12.00\%\scriptsize $+$1.5 & 11.87\%\scriptsize $+$1.1 & 12.05\%\scriptsize $+$1.4 & 19.68\%\scriptsize $+$1.9 \\
~ & SpanBERT & No & 9.90\%\scriptsize $+$1.0 & 12.50\%\scriptsize $+$1.2 & 6.00\%\scriptsize $+$2.0 & 12.80\%\scriptsize $+$1.3 & 13.00\%\scriptsize $+$1.7 & 12.60\%\scriptsize $+$1.5 & 19.15\%\scriptsize $+$2.0 \\
~ & WKLM $^\ast$ & Yes & 17.22\%\scriptsize $+$2.0 & 16.30\%\scriptsize $+$1.0 & 8.80\%\scriptsize $+$1.5 & 14.16\%\scriptsize $+$1.8 & 15.30\%\scriptsize $+$1.4 & 13.30\%\scriptsize $+$1.4 & 19.85\%\scriptsize $+$1.7 \\
~ & Splinter & No & 53.05\%\scriptsize $+$5.2 & 54.60\%\scriptsize $+$5.9 & 20.80\%\scriptsize $+$2.8 & 18.90\%\scriptsize $+$1.6 & 26.30\%\scriptsize $+$2.5 & 24.00\%\scriptsize $+$0.9 & 27.40\%\scriptsize $+$1.2 \\
\midrule
\multirow{4}*{PT} & RoBERTa $^\dagger$ & No & 39.50\%\scriptsize $+$1.1 & 27.10\%\scriptsize $+$2.0 & 12.20\%\scriptsize $+$3.9 & 16.82\%\scriptsize $+$2.0 & 19.10\%\scriptsize $+$1.8 & 22.26\%\scriptsize $+$1.9 & 20.18\%\scriptsize $+$2.2 \\
~ & P-tuning V2 & No & 60.48\%\scriptsize $+$4.2 & 59.10\%\scriptsize $+$4.4 & 22.33\%\scriptsize $+$2.9 & 22.42\%\scriptsize $+$0.7 & 28.08\%\scriptsize $+$4.1 & 26.33\%\scriptsize $+$2.3 & 27.52\%\scriptsize $+$2.4 \\
~ & \emph{\model} $_{\text{w/o. KPE}}$ & No & 63.07\%\scriptsize $+$3.6 & 64.22\%\scriptsize $+$4.3 & 23.80\%\scriptsize $+$2.0 & 21.35\%\scriptsize $+$0.8 & 29.41\%\scriptsize $+$3.1 & 27.80\%\scriptsize $+$2.6 & 27.95\%\scriptsize $+$2.4 \\ 
~ & \bf \emph{\model} & Yes & \bf 75.45\%\scriptsize $+$3.8 & \bf 67.05\%\scriptsize $+$4.7 & \bf 28.38\%\scriptsize $+$1.9 & \bf 24.80\%\scriptsize $+$2.4 & \bf 35.33\%\scriptsize $+$2.4 & \bf 33.90\%\scriptsize $+$2.0 & \bf 31.85\%\scriptsize $+$2.2 \\ 
\bottomrule
\end{tabular}
}
\end{small}
\caption{\label{tab:few-shot}
The averaged F1 performance of each benchmarks with standard deviation in few-shot scenario ($K=16$). FT and PT denote Fine-tuning and Prompt-tuning paradigms, respectively. RoBERTa$^\dagger$ in PT uses the vanilla MLM head to predict the answer text. WKLM $^\ast$ denotes our re-produced version based on RoBERTa-base.}
\label{tab:EQA}
\end{table*}

\subsection{Baselines}
To evaluate our proposed method, we consider the following methods as strong baselines: 1) \textbf{RoBERTa}~\cite{Liu19RoBERTa} is the optimized version of BERT, which introduces dynamic masking strategy. 2) \textbf{SpanBERT}~\cite{Joshi20SpanBERT} utilizes the span masking strategy and predicts the masked tokens based on boundary representations. 3) \textbf{WKLM}~\cite{Xiong2020Pretrained} belongs to knowledge-enhanced PLM, which continue to pre-trains on BERT with a novel entity replacement task. 4) \textbf{Splinter}~\cite{Ram2021Few} is the first work to regard span selection as a pre-training task for EQA. 5) \textbf{P-tuning-V2}~\cite{Liu21PTuningv2} is the prompt-based baseline for text generation tasks. 


    
    
    
    

\subsection{Benchmarks}
Our framework is evaluated over two benchmarks, including SQuAD2.0~\cite{Rajpurkar2018Know} and MRQA 2019 shared task~\cite{Fisch2019MRQA}.
The statistics of each dataset are shown in Appendix.

\noindent\textbf{SQuAD 2.0}~\cite{Rajpurkar2018Know}: It is a widely-used EQA benchmark, combining 43k unanswerable examples with original 87k answerable examples in SQuAD1.1~\cite{Rajpurkar2016SQuAD}. As the testing set is not publicly available, we use the public development set for the evaluation.
    
\noindent\textbf{MRQA 2019 shared task}~\cite{Fisch2019MRQA}: It is a shared task containing 6 EQA datasets formed in a unified format, such as
SQuAD1.1~\cite{Rajpurkar2016SQuAD}, NewsQA~\cite{trischler2017newsqa}, TriviaQA~\cite{Joshi17TriviaQA}, SearchQA~\cite{Dunn17SearchQA}, HotpotQA~\cite{Yang18HotpotQA} and NQ~\cite{Kwiatkowski2019Natural}.
Following~\cite{Ram2021Few}, we use the subset of Split I, where the training set is used for training and the development set is for evaluation.

\subsection{Implementation Details}

Follow the same settings as in~\cite{Ram2021Few}, for each EQA dataset, we randomly choose $K$ samples from the original training set to construct the few-shot training set and development set, respectively. As the test set is not available, we evaluate the model on the whole development set.



In our experiments, the underlying PLM is RoBERTa-base~\cite{Liu19RoBERTa} and the default hyper-parameters are initialized from the HuggingFace~\footnote{\url{https://huggingface.co/transformers/index.html}.}. We train our model by the Adam algorithm. The learning rate for MLM is fixed as 1e-5, while the initial learning rate for other new modules (self-attention in PPI) in \textit{\model} is set in \{1e-5, 3e-5, 5e-5, 1e-4\} with a warm-up rate of 0.1, the L2 weight decay 
value is $\gamma=0.01$. The balance hyper-parameter is set as $\lambda=0.5$. The number of \texttt{[MASK]} tokens in query prompts is $l_{mask}=10$. The number of negative spans is $S=5$. In few-shot settings, the definition scope of the sample number is $K\in\{16, 32, 64, \cdots, 512\}$. We set the batch size and the epoch number as 8 and 64, respectively.
During experiments, we choose five different random seeds $\{12, 21, 42, 87, 100\}$~\cite{Gao2021Making} and report the averaged performance. Because the generated answer text can be easy converted to a span with start and end position, we follow~\cite{Ram2021Few} to use the same F1 metric protocol, which measures the average overlap between the predicted and the ground-truth answer texts at the token level. 

\subsection{Main Results}

As shown in Table~\ref{tab:few-shot}, the results indicate that \textit{\model} outperforms all baselines with only 16 training examples. Surprisingly, we achieve 75.45\% and 67.05\% F1 values over SQuAD2.0~\cite{Rajpurkar2018Know} and SQuAD1.1~\cite{Rajpurkar2016SQuAD} with only 16 training examples, which outperforms the state-of-the-art method Splinter~\cite{Ram2021Few} by 22.40\% and 12.45\%, respectively. We also observe that the result of RoBERTa$^\dagger$ with vanilla MLM head is lower than any other of PT methods. It explains the necessity of the improvement of reasoning ability and the constraints on answer generation. To make fairly comparison, we also report the results of \emph{\model}$_{\text{w/o. KPE}}$, which is the basic model without injected KB. It makes a substantial improvement in all tasks, showing that prompt-tuning based on MLM generation is more suitable than span selection pre-training. 
In addition, we find that all results of traditional PLMs (e.g. RoBERTa~\cite{Liu19RoBERTa} and SpanBERT~\cite{Joshi20SpanBERT}) over seven tasks are lower than WKLM~\cite{Xiong2020Pretrained}, which injects domain-related knowledge into the PLM. Simultaneously, our model outperforms P-tuning V2~\cite{Liu21PTuningv2} and  \emph{\model}$_{\text{w/o. KPE}}$ by a large margin. These phenomenon indicate that EQA tasks can be further improved by injecting domain-related knowledge.

\begin{table}
\centering
\begin{small}
\resizebox{\linewidth}{!}{
\begin{tabular}{l|ccccc}
\toprule
\bf \#Training Samples$\longrightarrow$ & \bf 16 & \bf 1024 & \bf All \\
\midrule
\textbf{\textit{\model}}  & \textbf{75.45\%} & \textbf{84.90\%} & \textbf{90.85\%} \\
\midrule
w/o. KPE (w/o. PKI \& PPI) & 63.07\% & 73.17\% & 84.90\% \\
w/o. PPI & 73.36\% & 82.53\% & 90.70\% \\
w/o. SCL & 66.27\% & 74.40\% & 86.10\% \\
\bottomrule
\end{tabular}
}
\end{small}
\caption{The ablation F1 scores over SQuAD2.0 of \textit{\model} for few-shot learning setting. w/o. denotes that we only remove one component from \textit{\model}.}
\label{tab:ablation}
\end{table}




\begin{figure*}[t]
\vspace{-.25em}
\centering
\begin{tabular}{ccc}
\begin{minipage}[t]{0.3\linewidth}
    \includegraphics[width = 1\linewidth]{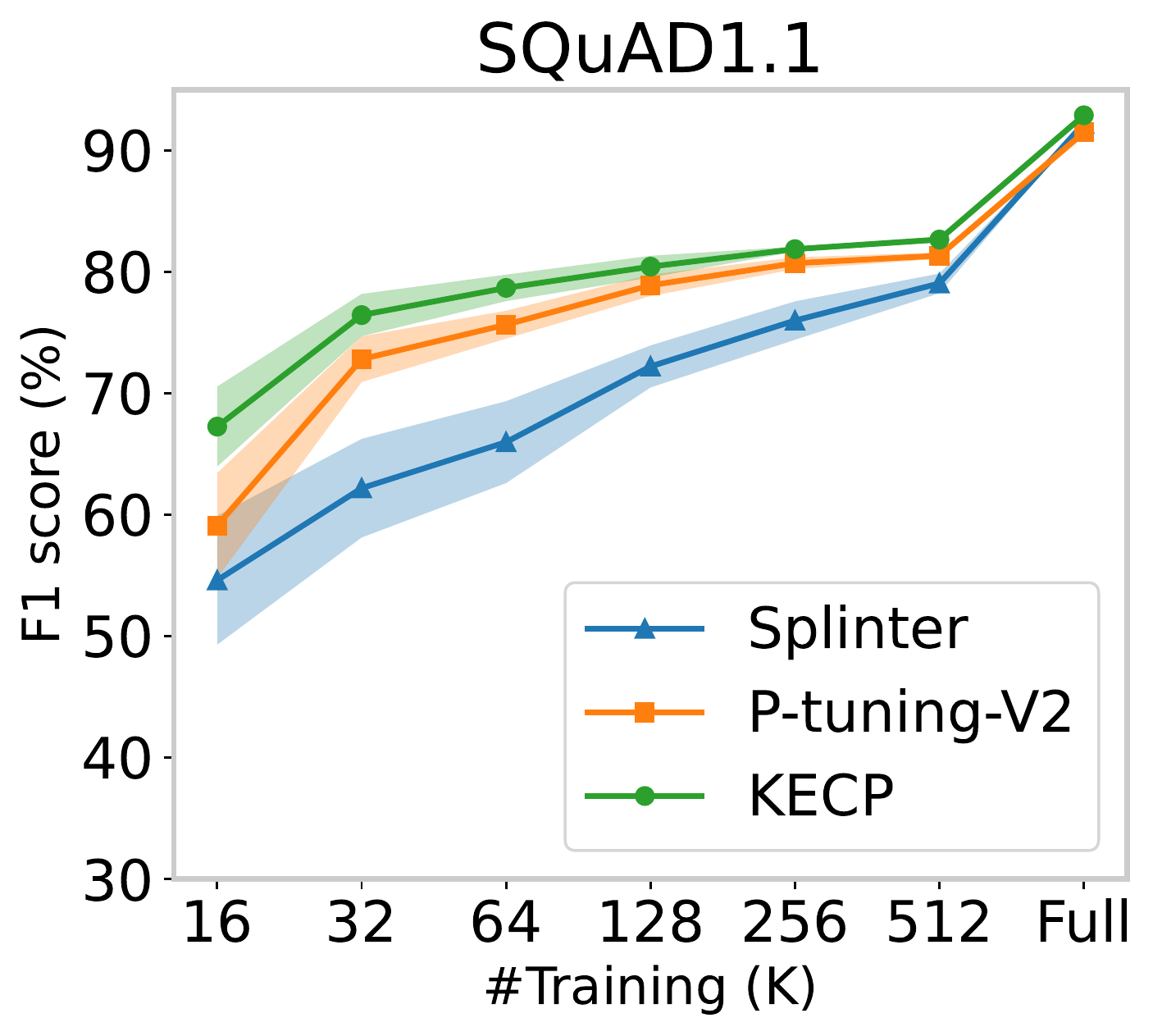}
\end{minipage}
\begin{minipage}[t]{0.3\linewidth}
    \includegraphics[width = 1\linewidth]{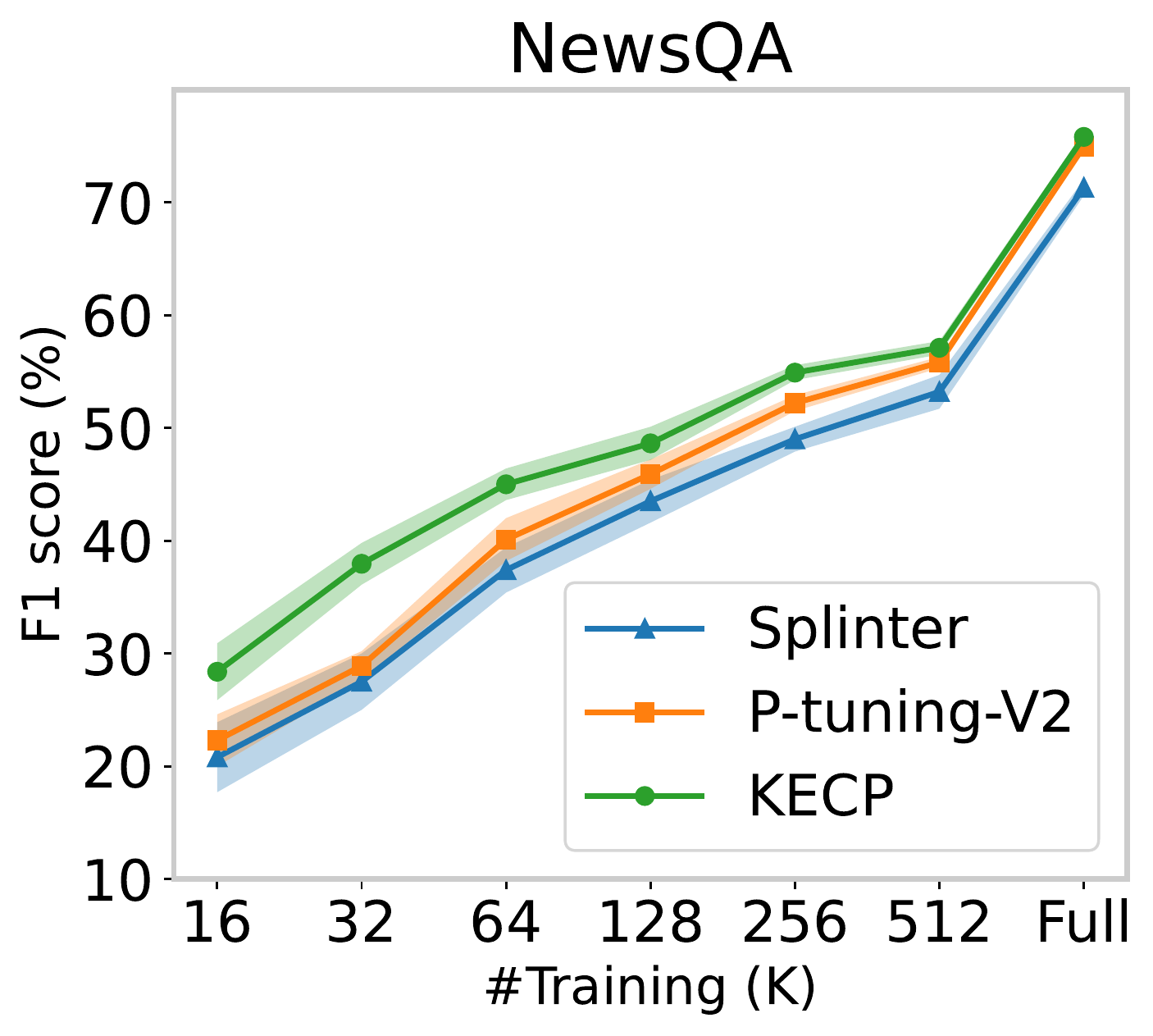}
\end{minipage}
\begin{minipage}[t]{0.3\linewidth}
    \includegraphics[width = 1\linewidth]{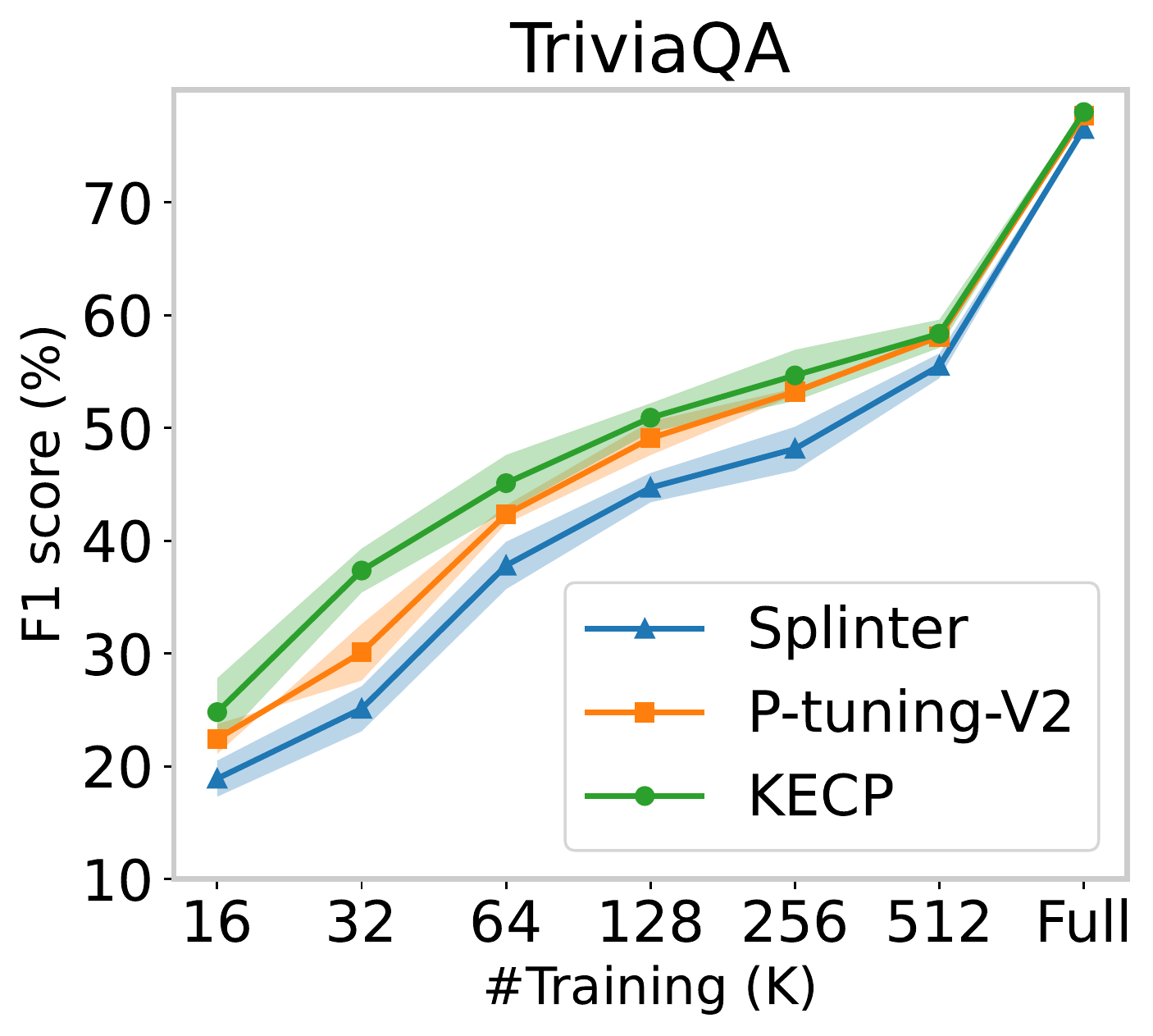}
\end{minipage}
\end{tabular}
\begin{tabular}{ccc}
\begin{minipage}[t]{0.3\linewidth}
    \includegraphics[width = 1\linewidth]{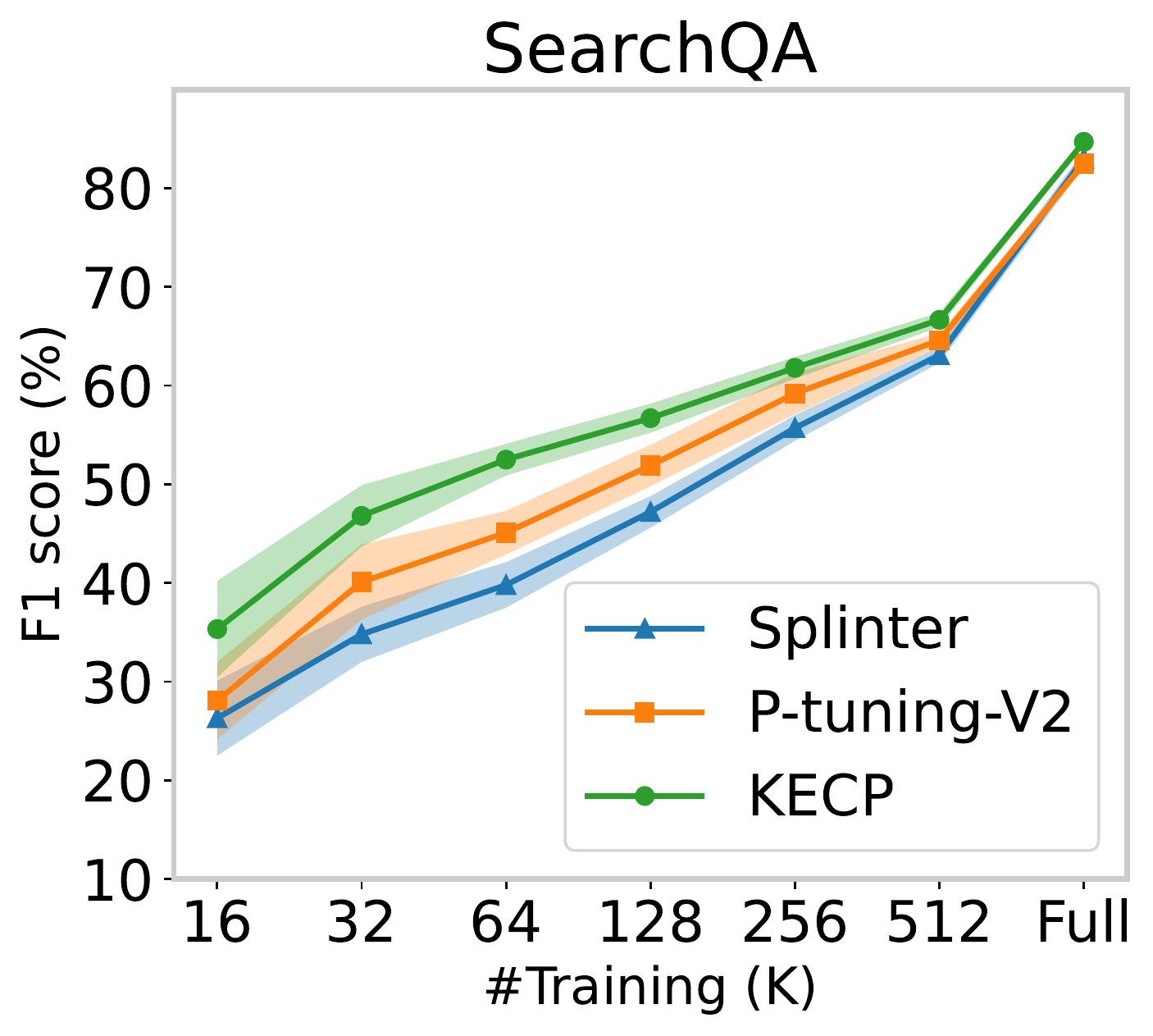}
\end{minipage}
\begin{minipage}[t]{0.3\linewidth}
    \includegraphics[width = 1\linewidth]{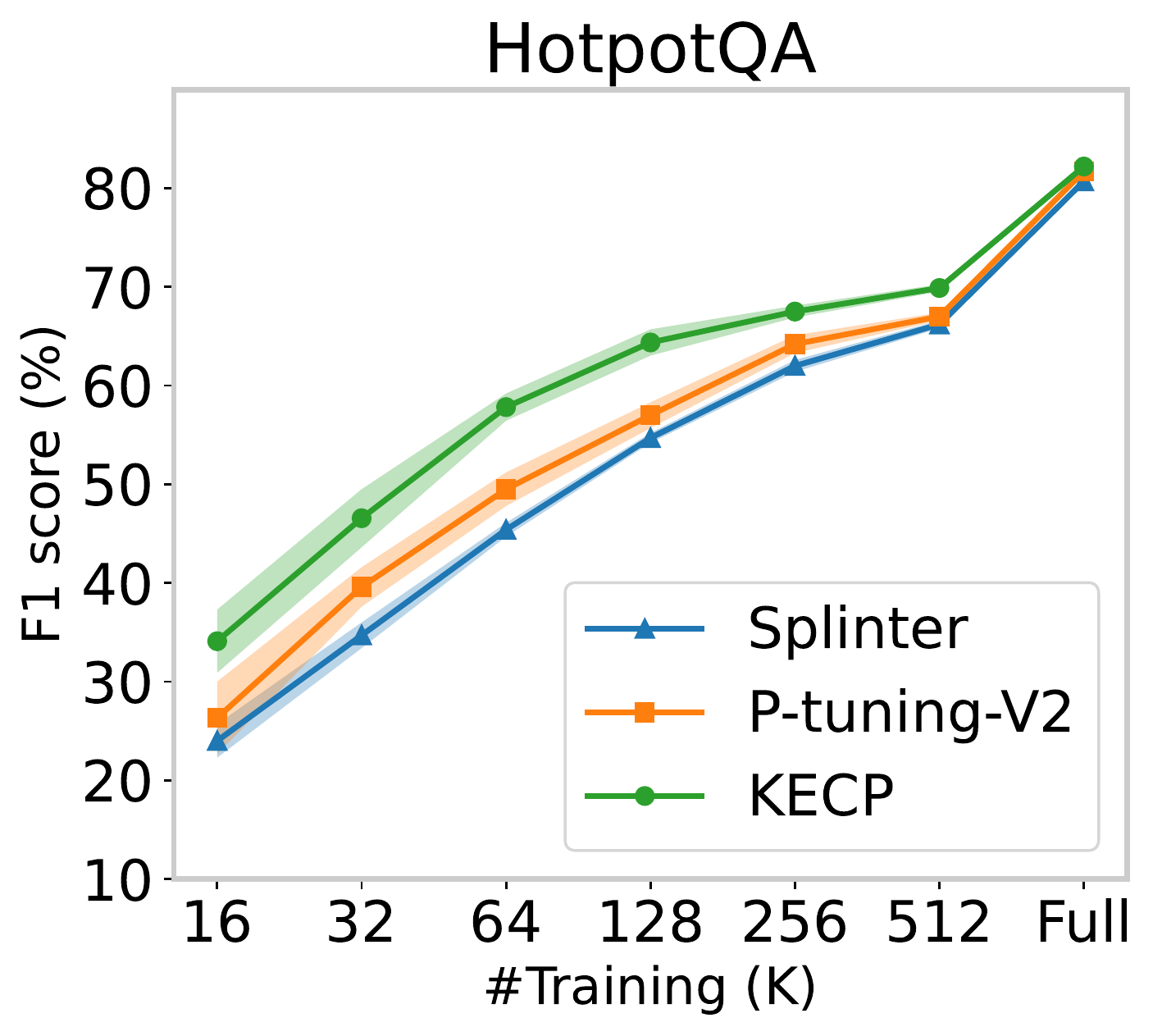}
\end{minipage}
\begin{minipage}[t]{0.3\linewidth}
    \includegraphics[width = 1\linewidth]{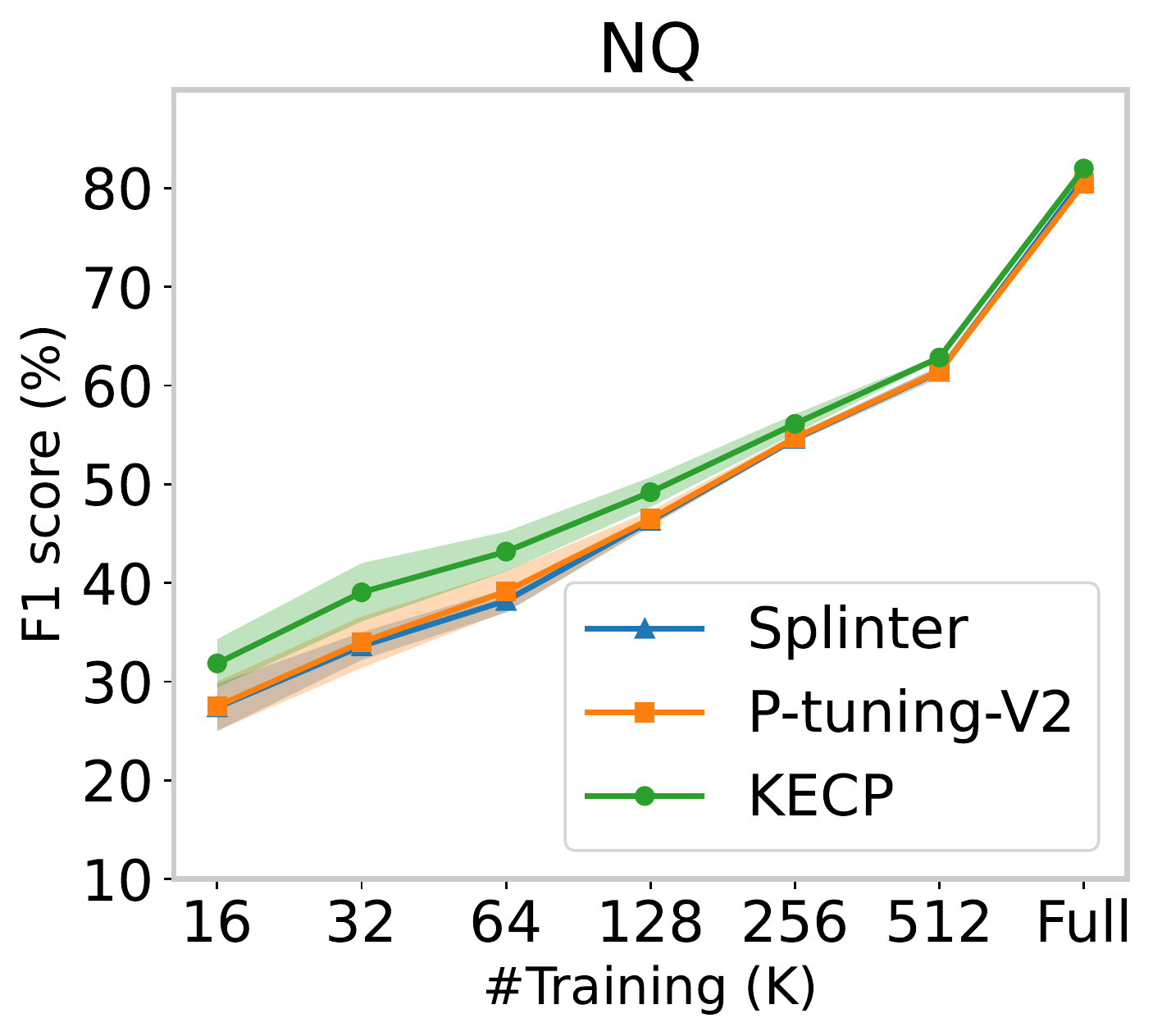}
\end{minipage}
\end{tabular}

\caption{Results of sample efficiency analysis. We compare~\emph{\model} with strong baselines with different numbers of training samples $K$ over MRQA 2019 shared tasks. ``Full'' denotes to the models trained over full training data.}
\label{fig:all_sample_effency}
\vspace{-.5em}
\end{figure*}

\begin{table}
\centering
\begin{small}
\resizebox{\linewidth}{!}{
\begin{tabular}[width=\columnwidth]{l|ccc}
\toprule
\textbf{Prompt Mapping} & \textbf{SQuAD2.0} & \textbf{NewsQA} & \textbf{HotpotQA} \\
\midrule
$\mathcal{T}_1$ (None) & 89.19\% & 72.15\% & 79.26\% \\
$\mathcal{T}_2$ (Manual) & 88.62\% & 72.70\% & 78.35\% \\
\midrule
$\mathcal{T}$ \textbf{(Proposed)} & \textbf{90.85\%} & \textbf{73.28\%} & \textbf{81.19\%} \\
\bottomrule
\end{tabular}
}
\end{small}
\caption{Comparison with proposed prompt template mapping $\mathcal{T}$ with two alternative methods $\mathcal{T}_1$ and $\mathcal{T}_2$.}
\label{tab:prompt}
\end{table}

\subsection{Detailed Analysis and Discussions}

\noindent\textbf{Ablation Study.}
To further understand why \textit{\model} achieves high performance, we perform an ablation analysis to better validate the contributions of each component. For simplicity, we only present the ablation experimental results on SQuAD2.0 with 16, 1024 and all training samples.

We show all ablation experiments in Table \ref{tab:ablation}, where w/o. KPE equals to the model without any domain-related knowledge (denotes to remove both PKI \& PPI). w/o. PPI denotes to only inject knowledge into selected prompt tokens without trainable self-attention. w/o. SCL means training without span-level contrastive learning (i.e. $\lambda=0$). We find that no matter which module is removed, the effect is decreasing. Particularly, when we remove both PKI and PPI, the performance is decreased by 12.38\%, 11.73\% and 5.95\%, respectively. The declines are larger than other cases, which indicates the significant impact of the passage-aware knowledge enhancement. We also find the SCL employed in this work also plays an important role in our framework, indicating that there are many confusing texts in the passage that need to be effectively distinguished by contrastive learning.

\noindent\textbf{Sample Efficiency.}
We further explore the model effects with different numbers $K$ of training samples. Figure \ref{fig:all_sample_effency} shows the performance with the different numbers of training samples over the MRQA 2019 shared task~\cite{Fisch2019MRQA}. Each point refers the averaged score across 5 randomly sampled datasets.
We observe that our~\emph{\model} consistently achieves higher scores regardless of the number of training samples. In particular, our method has more obvious advantages in low-resource scenarios than in full data settings. 
In addition, the results also indicate that prompt-tuning can be another novel paradigm for EQA.

\begin{table}[t]
\centering
\begin{small}
\resizebox{\linewidth}{!}{
\begin{tabular}{c|c|c|c|c}
\toprule
\textbf{Parameters} & \textbf{Values} & \textbf{Few} & \textbf{Full} & \textbf{Time} \\
\midrule
\multirow{4}*{
    \begin{tabular}{c}
        $l_{mask}=?$ \\
        $\lambda=0.5$ \\
        $S=5$
    \end{tabular}
}  & 4 & 39.20\% & 77.17\% & 0.9s \\
  & 7 & 41.35\% & 82.90\% & 1.3s \\
  & 10 & \textbf{42.30\%} & \textbf{83.27\%} & \textbf{1.5s} \\
  & 13 & 41.98\% & 82.84\% & 1.9s \\
\midrule
\multirow{5}*{
    \begin{tabular}{c}
        $\lambda=?$ \\
        $l_{mask}=10$ \\
        $S=5$
    \end{tabular}
} & 0 & 37.62\% & 76.91\% & 1.2s \\
  & 0.25 & 41.80\% & 82.99\% & 1.5s \\
  & 0.5 & \textbf{42.30\%} & \textbf{83.27\%} & \textbf{1.5s} \\
  & 0.75 & 42.09\% & 83.13\% & 1.5s \\
  & 1.0 & 40.10\% & 81.70\% & 1.6s \\
\midrule
\multirow{4}*{
    \begin{tabular}{c}
        $S=?$ \\
        $\lambda=0.5$ \\
        $l_{mask}=10$
    \end{tabular}
}  & 3 & 39.25\% & 80.02\% & 1.3s \\
  & 5 & \textbf{42.30\%} & \textbf{83.27\%} & \textbf{1.5s} \\
  & 7 & 42.30\% & 82.98\% & 1.9s \\
  & 9 & 42.41\% & 83.32\% & 2.3s \\
\bottomrule
\end{tabular}
}
\end{small}
\caption{The efficiency of hyper-parameters. All results are the average results of all datasets in both few-shot (Few) and full training data (Full) scenarios.}
\label{tab:hyper-parameter}
\end{table}

\noindent\textbf{Effects of Different Prompt Templates.}
In this part, we design two other template mappings:
\begin{itemize}
    \item $\mathcal{T}_1$ \textbf{(None)}: directly adding a series of \texttt{[MASK]} tokens without any template tokens.
    
    \item $\mathcal{T}_2$ \textbf{(Manual)}: designing a fixed template with multiple \texttt{[MASK]} tokens (e.g., ``The answer is \texttt{[MASK]}$\cdots$'').
\end{itemize}
To evaluate the efficiency of our proposed template mapping method compared with these baselines, we randomly select three tasks (i.e., SQuAD2.0, NewsQA and HotpotQA) and train models with full data. As shown in Table~\ref{tab:prompt}, we find that two simple templates have the similar performance. Our proposed method outperforms them by more than 1.0\% in terms of F1 score.~\footnote{We also provide intuitive cases in the experiments. More details can be found in the appendix.}


\noindent\textbf{Hyper-parameter Analysis.}
In this part, we investigate on some hyper-parameters in our framework, including the number of masked tokens $l_{mask}$, the balance coefficient $\lambda$ and the negative spans sampling number $S$. We also record the inference time over a batch with 8 testing examples. As shown in Table~\ref{tab:hyper-parameter}, when we tune $l_{mask}$, $\lambda$ and $S$ are fixed as 0.5 and 5, respectively. Results show that length of masked tokens plays an important role in prompt-tuning. We fix $l_{mask}=10, S=5$ and tune $\lambda$, and achieve the best performance when $\lambda=0.5$. We fix $\lambda=0.5, l_{mask}=10$ and tune the parameter $S$. We find the overall performance increases when increasing the sampled negatives. However, we recommend to set $S$ around 5 due to the faster inference speed.

\begin{figure}[t]
\centering
\includegraphics[width=\columnwidth]{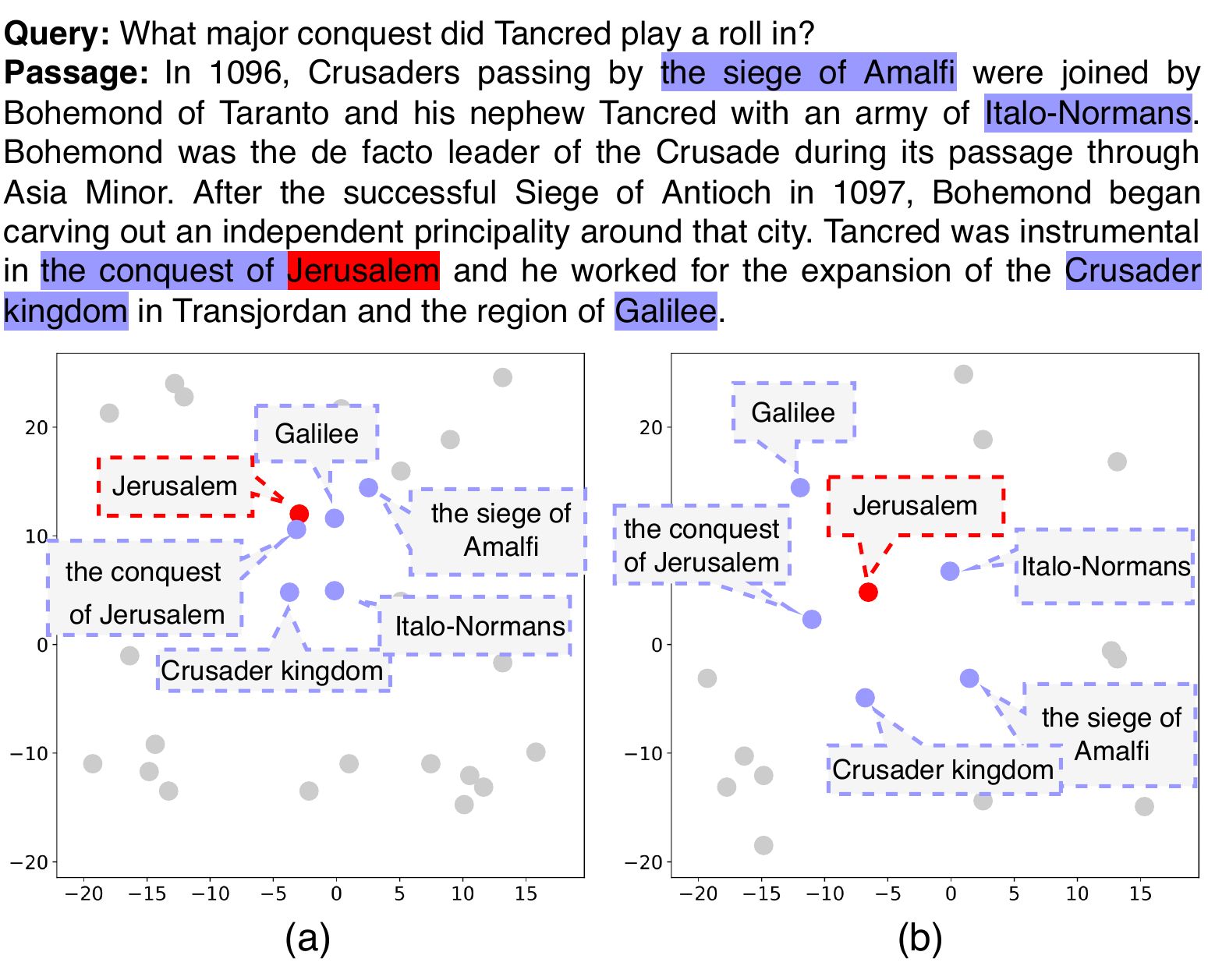} 
\caption{Visualizations of answer span texts. (a) is the result of the PLM without contrastive learning. (b) is the result of the PLM with contrastive learning.}
\label{fig: visual}
\end{figure}

\noindent\textbf{Effectiveness of Span-level Contrastive Learning.}
Furthermore, to evaluate how the model improved by span-level contrastive learning (SCL), we randomly select one example from the development set of SQuAD2.0~\cite{Rajpurkar2018Know}, and visualize it by t-sne~\cite{van2008visualizing} to gain more insight into the model performance. As shown in Figure~\ref{fig: visual}, the correct answer is ``Jerusalem'' (in red). We also obtain 5 negative spans (in blue) which may be confused with the correct answer. When the PLM is trained without SCL, in Figure~\ref{fig: visual}(a), we observe that all negative answers are agglomerated together with the correct answer ``Jerusalem''. It makes the PLM hard to search for the suitable results. In contrast, Figure~\ref{fig: visual}(b) represents the model trained with SCL. The result demonstrates that all negative spans can be better divided with the correct answer ``Jerusalem''. This shows that SCL in our \emph{\model} framework is reliable and can improve the performance for EQA. 


\noindent\textbf{The Accuracy of Answer Generation.}
A major difference between previous works and ours is that we model the EQA task as text generation.
Intuitively, if the model correctly generates the first answer token, it is easy to generate the remaining answer tokens because of the very small search space. Therefore, we analyze how difficult it is for the model to generate the first token correctly. 
Specifically, we check whether the generated first token and the first token of the ground truth are within a fixed window size $n_w$. 
As shown in Table~\ref{tab:window}, we find the accuracy of our method is lower than RoBERTa-base~\cite{Liu19RoBERTa} when $n_w=1$. Yet, we achieve the best performance when increasing the window size $n_w$ to 5. 
We think that our \emph{\model} can generate some rehabilitation text for the answer. For example in Figure~\ref{fig: visual}, the PLM may generate ``the conquest of Jerusalem'' rather than the correct answer with single token ``Jerusalem''. This phenomenon reflects the reason why we achieve lower accuracy when $n_w=1$. But, we think that the generated results are still in the vicinity of the correct answer.


\begin{table}
\centering
\begin{small}
\begin{tabular}{l|ccc}
\toprule
\textbf{Method} & \textbf{SQuAD2.0} & \textbf{NewsQA} & \textbf{HotpotQA} \\
\midrule
RoBERTa (\#1)  & 83.47\% & 69.80\% & 78.70\% \\
\midrule
\textit{\model} (\#1)  & 58.06\% & 51.30\% & 59.64\% \\
\textit{\model} (\#3)  & 74.57\% & 64.78\% & 72.11\% \\
\textbf{\textit{\model} (\#5)}  & \textbf{86.44\%} & \textbf{72.90\%} & \textbf{81.43\%} \\
\bottomrule
\end{tabular}
\end{small}
\caption{The accuracy of predicting the first \texttt{[MASK]} in the query prompt with full training samples for each task. \#$n_w$ denotes the window size.}
\label{tab:window}
\end{table}

\section{Conclusion}
To bridge the gap between the pre-training and fine-tuning objectives, \textit{\model} views EQA as an answer generation task. In \textit{\model}, the knowledge-aware prompt encoder injects external domain-related knowledge into the passage, and then enhances the representations of selected prompt tokens in the query.
The span-level contrastive learning objective is proposed to improve the performance of EQA.
Experiments on multiple benchmarks 
show that our framework outperforms the state-of-the-art methods. In the future, we will i) further improve the performance of \textit{\model} by applying controllable text generation techniques, and ii) explore the prompt-tuning for other types of MRC tasks, such as cloze-style MRC and multiple-choice MRC.

\appendix

\section{The Mapping Rules of Query Prompt}
\label{appendix:query_prompt}


Based on the analysis on the syntactic forms of queries from SQuAD~\cite{Rajpurkar2016SQuAD,Rajpurkar2018Know} and the MRQA 2019 shared task~\cite{Fisch2019MRQA}, we find that the queries in EQA can be directly transformed into the prompt templates with multiple \texttt{[MASK]} tokens. Let $\mathcal{T}:s\rightarrow s'$ be the prompt mapping where $s$ and $s'$ represent the original sentence and the prompt template, respectively. We list four rules for query prompt construction with corresponding example:
\begin{itemize}
    \item \textbf{Rule 1}. $\mathcal{T}$(\texttt{<s> be/done} $\cdots$ ?) $\rightarrow$ $\cdots$ \texttt{[MASK]} $\cdots$ \texttt{be/done} $\cdots$, where \texttt{<s>} can be chosen among \{"what", "who", "whose", "whom", "which", "how"\}.
    
    \item \textbf{Rule 2}. $\mathcal{T}$(\texttt{where be/done}$\cdots$?) $\rightarrow$ $\cdots$ \texttt{be/done at the place of [MASK]} $\cdots$.
    
    \item \textbf{Rule 3}. $\mathcal{T}$(\texttt{when be/done} $\cdots$?) $\rightarrow$  $\cdots$ \texttt{be/done at the time of [MASK]} $\cdots$.
    
    
    \item \textbf{Rule 4}. $\mathcal{T}$(\texttt{why be/done} $\cdots$?) $\rightarrow$ \texttt{the reason why} $\cdots$ \texttt{be/done [MASK]} $\cdots$.
\end{itemize}
For the query that does not match these rules will be directly appended with multiple masked language tokens. Table~\ref{tab:mapping_rule} shows the examples of each mapping rule.

\begin{table*}
\centering
\begin{small}
\begin{tabular}{p{0.08\textwidth}|p{0.38\textwidth}|p{0.42\textwidth}}
\toprule
\bf Rule & \bf Original Query & \bf Query Prompt \\
\midrule
Rule 1 & A Japanese manga series based on a 16 year old high school student Ichitaka Seto, is written and illustrated by someone born in what year? & A Japanese manga series based on a 16 year old high school student Ichitaka Seto, is written and illustrated by someone born in \texttt{[MASK]} \texttt{[MASK]}.\\
\midrule
Rule 2 & 
Where is the company that Sachin Warrier worked for as a software engineer? & The company that Sachin Warrier worked for as a software engineer is at the place of \texttt{[MASK]} \texttt{[MASK]}. \\
\midrule
Rule 3 & When the Canberra was introduced to service with the Royal Air Force (RAF), the type's first operator, in May 1951, it became the service's first jet-powered bomber aircraft. & The Canberra was introduced to service with the Royal Air Force (RAF) at the time of \texttt{[MASK]} \texttt{[MASK]}, the type's first operator, in May 1951, it became the service's first jet-powered bomber aircraft. \\
\midrule
Rule 4 & Why did Rudolf Hess stop serving Hitler in 1941? & The reason why did Rudolf Hess stop serving Hitler in 1941 is that \texttt{[MASK]} \texttt{[MASK]}. \\
\midrule
Other & How much longer after he was born did Werder Bremen get founded in the northwest German federal state Free Hanseatic City of Bremen? & How much longer after he was born did Werder Bremen get founded in the northwest German federal state Free Hanseatic City of Bremen? \texttt{[MASK]} \texttt{[MASK]}. \\
\midrule
\end{tabular}
\end{small}
\caption{Example of each query prompt mapping rule.}
\label{tab:mapping_rule}
\end{table*}

\section{Data Sources}

In this section, we give more details on data sources used in the experiments.

\subsection{The Benchmarks of EQA}
\label{appendix:benchmarks}

\begin{table}
\centering
\begin{small}
\begin{tabular}{c|cccc}
\toprule
\bf Dataset & \bf \#Train & \bf \#Dev & \bf \#All \\
\midrule
SQuAD2.0 & 118,446 & 11,873 & 130,319 \\
\midrule
SQuAD1.1 & 86,588 & 10,507 & 97,095 \\
NewsQA & 74,160 & 4,212 & 78,372 \\
TriviaQA & 61,688 & 7,785 & 69,573 \\
SearchQA & 117,384 & 16,980 & 134,364 \\
HotpotQA & 72,928 & 5,904 & 78,832 \\
NQ & 104,071 & 12,836 & 116,907 \\
\midrule
\end{tabular}
\end{small}
\caption{The statistics of multiple EQA benchmarks.}
\label{tab:data}
\end{table}

We choose two widely used EQA benchmarks for the evaluation, including SQuAD2.0~\cite{Rajpurkar2018Know} and the MRQA 2019 shared task~\cite{Fisch2019MRQA}. Specifically, the MRQA 2019 shared task was proposed to evaluate the domain transferable of neural models, where the authors selected 18 distinct question answering datasets, then adapted and unified them into the same format. They divided all datasets into 3 splits, where Split I is used for model training and development, Split II is used for development only and Split III is used for evaluation. Because our work focuses on few-shot learning settings, we simply choose 6 dataset from Split I in our experiments, including SQuAD1.1~\cite{Rajpurkar2016SQuAD}, NewsQA~\cite{trischler2017newsqa}, TriviaQA~\cite{Joshi17TriviaQA}, SearchQA~\cite{Dunn17SearchQA}, HotpotQA~\cite{Yang18HotpotQA} and NQ~\cite{Kwiatkowski2019Natural}. We also choose SQuAD2.0~\cite{Rajpurkar2018Know} to conduct evaluations.

In few-shot learning settings, for each dataset, we randomly select $K$ examples with five different random seeds for training and development, respectively. For the full data settings, we follow the same settings of Splinter~\cite{Ram2021Few} to use all training data.

    


\subsection{External Knowledge Base}
\label{appendix: kb}

For the domain-related knowledge base, we use WikiData5M~\cite{Wang21KEPLER}, which is a large-scale knowledge graph aligned with text descriptions from the corresponding Wikipedia pages. It consists of 4,594,485 entities, 20,510,107 triples and 822 relation types. We use the ConVE~\cite{Tim18Convolutional} algorithm to pre-train the entity and relation embeddings. We set its dimension as 512, the negative sampling size as 64, the batch size as 128 and the learning rate as 0.001. Finally, we only store the embeddings of all the entities. For the passage knowledge injection, we use entity linking tools (e.g, TAGME tool in python~\footnote{https://pypi.org/project/tagme/.}) to align the entity mentions in passages. The embeddings of tokens are calculated by the lemmatization operator~\cite{Dai2021Incorporating}.

\section{Details of Negative Span Sampling}
\label{appendix:csc}

In order to construct negative spans for span-level contrastive learning (SCL), we follow a simple pipeline to implement confusion span sampling. At first, we use slide window to obtain a series of span texts. Next, we filter out span texts which are incomplete sequences or dissatisfy the lexical and grammatical rules. Finally, we calculate the semantic similarity between each candidate span text and the true answer. Formally, suppose $Y=y_1, y_2, \cdots, y_l$ is the ground truth. Given one candidate span $X=x_1, x_2, \cdots, y_{l'}$, where $l, l'$ are the lengths of the ground truth and the candidate span text, respectively, we have:
\begin{equation}
\begin{aligned}
\text{Sim}(X, Y) = & \texttt{dist}(X, Y) \\
& \cdot\texttt{cos}(\frac{1}{l}\sum_{i=1}^{l}\mathbf{y}_i, \frac{1}{l'}\sum_{i=1}^{l'}\mathbf{y}'_i)
\label{eqn:sim-cos}
\end{aligned}
\end{equation}
where $\mathbf{y}_i$, $\mathbf{y}'_i$ denote the knowledge-injected representations of $i$-th token, respectively. $\texttt{cos}(X, Y)$ aims to compute the cosine similarity between $X$ and $Y$. We also introduce the $\texttt{dist}(X, Y)$ function to represent the normalized position distance between $X$ and $Y$ by the intuition that the text closer to the correct answer is prone to confusion. Specifically, for each candidate $X$, we obtain the distance between the first token of $X$ and $Y$, and calculate the normalized weight for each candidate. For example in Figure~\ref{fig1}, the distance between the candidate ``avid Crusad'' and the answer ``fighting horsemen'' is 16, and the normalized weight is 0.15.


\begin{table}
\centering
\begin{small}
\begin{tabular}{l|ccccc}
\toprule
\bf \#Training Samples$\longrightarrow$ & \bf 16 & \bf 1024 & \bf All \\
\midrule
\textbf{\textit{\model}}  & \textbf{75.45\%} & \textbf{84.90\%} & \textbf{90.85\%} \\
\midrule
w/o. SCL & 66.27\% & 74.40\% & 86.10\% \\
w/o. filter \& sort & 71.35\% & 79.05\% & 87.80\% \\
w/o. \texttt{dist} & 74.90\% & 84.60\% & 90.55\% \\
\bottomrule
\end{tabular}
\end{small}
\caption{The ablation F1 scores over SQuAD2.0 to evaluate the importance of each technique in the confusion span contrastive task. w/o. denotes that we only remove one component from \textit{\model}.}
\label{tab:csc_ablation}
\end{table}

We provide a brief ablation study for this module. Specifically, w/o. SCL means that we remove all techniques of this module (setting $\lambda=0$ in Equation~(\ref{eqn:loss})). w/o. filter \& sort denotes randomly sampling $S$ spans without the pipeline. w/o. \texttt{dist} represents setting $\texttt{dist}(X, Y)=1$ in Equation~(\ref{eqn:sim-cos}).
As shown in Table~\ref{tab:csc_ablation}, the results demonstrate that our model can be improved by the combination of all techniques.


\begin{thebibliography}{37}
    \expandafter\ifx\csname natexlab\endcsname\relax\def\natexlab#1{#1}\fi
    
    \bibitem[{Brown et~al.(2020)Brown, Mann, and Nick~Ryder}]{Brown2020Language}
    Tom~B. Brown, Benjamin Mann, and etc. Nick~Ryder. 2020.
    \newblock Language models are few-shot learners.
    \newblock In \emph{NeurIPS}.
    
    \bibitem[{Chen et~al.(2020)Chen, Kornblith, Norouzi, and Hinton}]{Chen2020a}
    Ting Chen, Simon Kornblith, Mohammad Norouzi, and Geoffrey~E. Hinton. 2020.
    \newblock A simple framework for contrastive learning of visual
      representations.
    \newblock In \emph{{ICML}}, volume 119, pages 1597--1607.
    
    \bibitem[{Dai et~al.(2021)Dai, Zheng, Sui, and Chang}]{Dai2021Incorporating}
    Damai Dai, Hua Zheng, Zhifang Sui, and Baobao Chang. 2021.
    \newblock Incorporating connections beyond knowledge embeddings: {A}
      plug-and-play module to enhance commonsense reasoning in machine reading
      comprehension.
    \newblock \emph{CoRR}, abs/2103.14443.
    
    \bibitem[{Dettmers et~al.(2018)Dettmers, Minervini, Stenetorp, and
      Riedel}]{Tim18Convolutional}
    Tim Dettmers, Pasquale Minervini, Pontus Stenetorp, and Sebastian Riedel. 2018.
    \newblock Convolutional 2d knowledge graph embeddings.
    \newblock In \emph{AAAI}.
    
    \bibitem[{Devlin et~al.(2019)Devlin, Chang, Lee, and
      Toutanova}]{Devlin2019BERT}
    Jacob Devlin, Ming{-}Wei Chang, Kenton Lee, and Kristina Toutanova. 2019.
    \newblock {BERT:} pre-training of deep bidirectional transformers for language
      understanding.
    \newblock In \emph{NAACL-HLT}, pages 4171--4186.
    
    \bibitem[{Dunn et~al.(2017)Dunn, Sagun, Higgins, G{\"{u}}ney, Cirik, and
      Cho}]{Dunn17SearchQA}
    Matthew Dunn, Levent Sagun, Mike Higgins, V.~Ugur G{\"{u}}ney, Volkan Cirik,
      and Kyunghyun Cho. 2017.
    \newblock Searchqa: {A} new q{\&}a dataset augmented with context from a search
      engine.
    \newblock \emph{CoRR}, abs/1704.05179.
    
    \bibitem[{Fisch et~al.(2019)Fisch, Talmor, Jia, Seo, Choi, and
      Chen}]{Fisch2019MRQA}
    Adam Fisch, Alon Talmor, Robin Jia, Minjoon Seo, Eunsol Choi, and Danqi Chen.
      2019.
    \newblock {MRQA} 2019 shared task: Evaluating generalization in reading
      comprehension.
    \newblock In \emph{EMNLP}, pages 1--13.
    
    \bibitem[{Gao et~al.(2021)Gao, Fisch, and Chen}]{Gao2021Making}
    Tianyu Gao, Adam Fisch, and Danqi Chen. 2021.
    \newblock Making pre-trained language models better few-shot learners.
    \newblock In \emph{ACL}, pages 3816--3830.
    
    \bibitem[{Han et~al.(2021)Han, Zhao, Ding, Liu, and Sun}]{Xu2021PTR}
    Xu~Han, Weilin Zhao, Ning Ding, Zhiyuan Liu, and Maosong Sun. 2021.
    \newblock {PTR:} prompt tuning with rules for text classification.
    \newblock \emph{CoRR}, abs/2105.11259.
    
    \bibitem[{Joshi et~al.(2020)Joshi, Chen, Liu, Weld, Zettlemoyer, and
      Levy}]{Joshi20SpanBERT}
    Mandar Joshi, Danqi Chen, Yinhan Liu, Daniel~S. Weld, Luke Zettlemoyer, and
      Omer Levy. 2020.
    \newblock Spanbert: Improving pre-training by representing and predicting
      spans.
    \newblock \emph{TACL}, 64--77.
    
    \bibitem[{Joshi et~al.(2017)Joshi, Choi, Weld, and
      Zettlemoyer}]{Joshi17TriviaQA}
    Mandar Joshi, Eunsol Choi, Daniel~S. Weld, and Luke Zettlemoyer. 2017.
    \newblock Triviaqa: {A} large scale distantly supervised challenge dataset for
      reading comprehension.
    \newblock In \emph{ACL}, pages 1601--1611.
    
    \bibitem[{Kwiatkowski et~al.(2019)Kwiatkowski, Palomaki, Redfield, and
      et~al.}]{Kwiatkowski2019Natural}
    Tom Kwiatkowski, Jennimaria Palomaki, Olivia Redfield, and et~al. 2019.
    \newblock Natural questions: a benchmark for question answering research.
    \newblock \emph{TACL}.
    
    \bibitem[{Lai et~al.(2017)Lai, Xie, Liu, Yang, and Hovy}]{Lai2017RACE}
    Guokun Lai, Qizhe Xie, Hanxiao Liu, Yiming Yang, and Eduard~H. Hovy. 2017.
    \newblock {RACE:} large-scale reading comprehension dataset from examinations.
    \newblock In \emph{EMNLP}, pages 785--794.
    
    \bibitem[{Levy et~al.(2017)Levy, Seo, Choi, and Zettlemoyer}]{Levy2017Zero}
    Omer Levy, Minjoon Seo, Eunsol Choi, and Luke Zettlemoyer. 2017.
    \newblock Zero-shot relation extraction via reading comprehension.
    \newblock In \emph{CoNLL}, pages 333--342.
    
    \bibitem[{Li and Liang(2021{\natexlab{a}})}]{Xiang2021Prefix}
    Xiang~Lisa Li and Percy Liang. 2021{\natexlab{a}}.
    \newblock Prefix-tuning: Optimizing continuous prompts for generation.
    \newblock In \emph{ACL/IJCNLP}, pages 4582--4597. Association for Computational
      Linguistics.
    
    \bibitem[{Li and Liang(2021{\natexlab{b}})}]{Li2021Prefix}
    Xiang~Lisa Li and Percy Liang. 2021{\natexlab{b}}.
    \newblock Prefix-tuning: Optimizing continuous prompts for generation.
    \newblock In \emph{ACL}, pages 4582--4597.
    
    \bibitem[{Liu et~al.(2021{\natexlab{a}})Liu, Ji, Fu, Du, Yang, and
      Tang}]{Liu21PTuningv2}
    Xiao Liu, Kaixuan Ji, Yicheng Fu, Zhengxiao Du, Zhilin Yang, and Jie Tang.
      2021{\natexlab{a}}.
    \newblock P-tuning v2: Prompt tuning can be comparable to fine-tuning
      universally across scales and tasks.
    \newblock \emph{CoRR}.
    
    \bibitem[{Liu et~al.(2021{\natexlab{b}})Liu, Zheng, Du, Ding, Qian, Yang, and
      Tang}]{Xiao2021GPT}
    Xiao Liu, Yanan Zheng, Zhengxiao Du, Ming Ding, Yujie Qian, Zhilin Yang, and
      Jie Tang. 2021{\natexlab{b}}.
    \newblock {GPT} understands, too.
    \newblock \emph{CoRR}, abs/2103.10385.
    
    \bibitem[{Liu et~al.(2019)Liu, Ott, Goyal, Du, Joshi, Chen, and
      et~al.}]{Liu19RoBERTa}
    Yinhan Liu, Myle Ott, Naman Goyal, Jingfei Du, Mandar Joshi, Danqi Chen, and
      et~al. 2019.
    \newblock Roberta: {A} robustly optimized {BERT} pretraining approach.
    \newblock \emph{CoRR}.
    
    \bibitem[{Qin and Eisner(2021)}]{Qin2021Learning}
    Guanghui Qin and Jason Eisner. 2021.
    \newblock Learning how to ask: Querying lms with mixtures of soft prompts.
    \newblock In \emph{NAACL-HLT}, pages 5203--5212.
    
    \bibitem[{Rajpurkar et~al.(2018)Rajpurkar, Jia, and Liang}]{Rajpurkar2018Know}
    Pranav Rajpurkar, Robin Jia, and Percy Liang. 2018.
    \newblock Know what you don't know: Unanswerable questions for squad.
    \newblock \emph{CoRR}, abs/1806.03822.
    
    \bibitem[{Rajpurkar et~al.(2016)Rajpurkar, Zhang, Lopyrev, and
      Liang}]{Rajpurkar2016SQuAD}
    Pranav Rajpurkar, Jian Zhang, Konstantin Lopyrev, and Percy Liang. 2016.
    \newblock Squad: 100, 000+ questions for machine comprehension of text.
    \newblock \emph{CoRR}.
    
    \bibitem[{Ram et~al.(2021)Ram, Kirstain, Berant, Globerson, and
      Levy}]{Ram2021Few}
    Ori Ram, Yuval Kirstain, Jonathan Berant, Amir Globerson, and Omer Levy. 2021.
    \newblock Few-shot question answering by pretraining span selection.
    \newblock In \emph{ACL}.
    
    \bibitem[{Schick and Sch{\"{u}}tze(2021)}]{Timo2021Exploiting}
    Timo Schick and Hinrich Sch{\"{u}}tze. 2021.
    \newblock Exploiting cloze-questions for few-shot text classification and
      natural language inference.
    \newblock In \emph{EACL}, pages 255--269.
    
    \bibitem[{Shin et~al.(2020)Shin, Razeghi, IV, Wallace, and
      Singh}]{Shin2020AutoPrompt}
    Taylor Shin, Yasaman Razeghi, Robert L.~Logan IV, Eric Wallace, and Sameer
      Singh. 2020.
    \newblock Autoprompt: Eliciting knowledge from language models with
      automatically generated prompts.
    \newblock In \emph{EMNLP}.
    
    \bibitem[{Trischler et~al.(2017)Trischler, Wang, Yuan, Harris, Sordoni,
      Bachman, and Suleman}]{trischler2017newsqa}
    Adam Trischler, Tong Wang, Xingdi Yuan, Justin Harris, Alessandro Sordoni,
      Philip Bachman, and Kaheer Suleman. 2017.
    \newblock {N}ews{QA}: A machine comprehension dataset.
    \newblock In \emph{WRLNLP}, pages 191--200.
    
    \bibitem[{Van~der Maaten and Hinton(2008)}]{van2008visualizing}
    Laurens Van~der Maaten and Geoffrey Hinton. 2008.
    \newblock Visualizing data using t-sne.
    \newblock \emph{Journal of machine learning research}, 9(11).
    
    \bibitem[{Vinyals et~al.(2015)Vinyals, Fortunato, and
      Jaitly}]{Vinyals15Pointer}
    Oriol Vinyals, Meire Fortunato, and Navdeep Jaitly. 2015.
    \newblock Pointer networks.
    \newblock In \emph{{NIPS}}, pages 2692--2700.
    
    \bibitem[{Wang and Jiang(2019)}]{Wang2019Explicit}
    Chao Wang and Hui Jiang. 2019.
    \newblock Explicit utilization of general knowledge in machine reading
      comprehension.
    \newblock In \emph{ACL}, pages 2263--2272. Association for Computational
      Linguistics.
    
    \bibitem[{Wang et~al.(2022)Wang, Qiu, Zhang, Liu, Li, Wang, Wang, Huang, and
      Lin}]{DBLP:journals/corr/abs-2205-00258}
    Chengyu Wang, Minghui Qiu, Taolin Zhang, Tingting Liu, Lei Li, Jianing Wang,
      Ming Wang, Jun Huang, and Wei Lin. 2022.
    \newblock Easynlp: {A} comprehensive and easy-to-use toolkit for natural
      language processing.
    \newblock \emph{CoRR}, abs/2205.00258.
    
    \bibitem[{Wang and Jiang(2017)}]{Wang2017Machine}
    Shuohang Wang and Jing Jiang. 2017.
    \newblock Machine comprehension using match-lstm and answer pointer.
    \newblock In \emph{ICLR}.
    
    \bibitem[{Wang et~al.(2017)Wang, Yang, Wei, Chang, and Zhou}]{wang2017gated}
    Wenhui Wang, Nan Yang, Furu Wei, Baobao Chang, and Ming Zhou. 2017.
    \newblock Gated self-matching networks for reading comprehension and question
      answering.
    \newblock In \emph{ACL}, pages 189--198.
    
    \bibitem[{Wang et~al.(2021)Wang, Gao, Zhu, Zhang, Liu, Li, and
      Tang}]{Wang21KEPLER}
    Xiaozhi Wang, Tianyu Gao, Zhaocheng Zhu, Zhengyan Zhang, Zhiyuan Liu, Juanzi
      Li, and Jian Tang. 2021.
    \newblock {KEPLER:} {A} unified model for knowledge embedding and pre-trained
      language representation.
    \newblock \emph{TACL}, 9:176--194.
    
    \bibitem[{Xiong et~al.(2020)Xiong, Du, Wang, and
      Stoyanov}]{Xiong2020Pretrained}
    Wenhan Xiong, Jingfei Du, William~Yang Wang, and Veselin Stoyanov. 2020.
    \newblock Pretrained encyclopedia: Weakly supervised knowledge-pretrained
      language model.
    \newblock In \emph{{ICLR}}.
    
    \bibitem[{Yang et~al.(2019)Yang, Wang, Liu, Liu, Lyu, Wu, She, and
      Li}]{Yang2019Enhancing}
    An~Yang, Quan Wang, Jing Liu, Kai Liu, Yajuan Lyu, Hua Wu, Qiaoqiao She, and
      Sujian Li. 2019.
    \newblock Enhancing pre-trained language representations with rich knowledge
      for machine reading comprehension.
    \newblock In \emph{ACL}, pages 2346--2357.
    
    \bibitem[{Yang et~al.(2018)Yang, Qi, Zhang, Bengio, Cohen, Salakhutdinov, and
      Manning}]{Yang18HotpotQA}
    Zhilin Yang, Peng Qi, Saizheng Zhang, Yoshua Bengio, William~W. Cohen, Ruslan
      Salakhutdinov, and Christopher~D. Manning. 2018.
    \newblock Hotpotqa: {A} dataset for diverse, explainable multi-hop question
      answering.
    \newblock In \emph{{EMNLP}}, pages 2369--2380.
    
    \bibitem[{Zhang et~al.(2021)Zhang, Deng, Cheng, Chen, Zhang, Zhang, Chen, and
      Center}]{zhang2021drop}
    Ningyu Zhang, Shumin Deng, Xu~Cheng, Xi~Chen, Yichi Zhang, Wei Zhang, Huajun
      Chen, and Hangzhou~Innovation Center. 2021.
    \newblock Drop redundant, shrink irrelevant: Selective knowledge injection for
      language pretraining.
    \newblock In \emph{IJCAI}.
    
    \end{thebibliography}
\end{document}